\documentclass[sigconf]{acmart}

\usepackage{colortbl}
\definecolor{mygray}{gray}{.9}
\usepackage{multirow}
\usepackage{pifont}
\usepackage{subfigure}
\usepackage{tikz}   
\usepackage{booktabs}
\usepackage{colortbl}
\usepackage{makecell}
\usepackage[capitalize]{cleveref}
\usepackage{subcaption}
\usepackage{amsmath}
\usepackage{hyperref}
\hypersetup{
    colorlinks=true,
    linkcolor=blue,
    filecolor=magenta,      
    urlcolor=blue,
}
\crefname{section}{Sec.}{Secs.}
\Crefname{section}{Section}{Sections}
\Crefname{table}{Table}{Tables}
\crefname{table}{Tab.}{Tabs.}
\Crefname{equation}{Equation}{Equations}
\crefname{equation}{Eq.}{Eqs.}

\newcommand\eg{\emph{e.g.}} 
\newcommand\ie{\emph{i.e.}}

\newcommand{\vI}{\mathbf{I}}
\newcommand{\vA}{\mathbf{A}}
\newcommand{\vJ}{\mathbf{J}}
\newcommand{\vt}{\mathbf{T}}
\newcommand{\vd}{\mathbf{D}}
\newcommand{\cG}{\mathcal{G}}

\definecolor{myGreen}{RGB}{0, 176, 80}
\definecolor{lightGreen}{RGB}{128,158,108}
\definecolor{lightBlue}{RGB}{149,175,198}
\AtBeginDocument{%
  }

\acmISBN{978-1-4503-XXXX-X/18/06}

\renewcommand\footnotetextcopyrightpermission[1]{}

\begin{document}


\title{LMHaze: Intensity-aware Image Dehazing with a Large-scale Multi-intensity Real Haze Dataset}

\author{Ruikun Zhang}
\orcid{0009-0008-3104-2218}
\affiliation{%
  \institution{Beijing Institute of Technology, Beijing, China}
  \city{}
  \country{}
  }
\email{ruikun.zhang@bit.edu.cn}

\author{Hao Yang}
\affiliation{%
  \institution{Beijing Institute of Technology, Beijing, China}
  \city{}
  \country{}
  }
\email{hao.yang@bit.edu.cn}

\author{Yan Yang}
\affiliation{%
  \institution{Australian National University, Canberra, Australia}
  \city{}
  \country{}
  }
\email{u6169130@anu.edu.au}

\author{Ying Fu}
\affiliation{%
  \institution{Beijing Institute of Technology, Beijing, China}
  \city{}
  \country{}
  }
\email{fuying@bit.edu.cn}

\author{Liyuan Pan}
\authornote{Corresponding author.}
\affiliation{%
  \institution{Beijing Institute of Technology, Beijing, China}
  \city{}
  \country{}
  }
\email{liyuan.pan@bit.edu.cn}


\begin{abstract}


Image dehazing has drawn a significant attention in recent years. Learning-based methods usually require paired hazy and corresponding ground truth (haze-free) images for training. However, it is difficult to collect real-world image pairs, which prevents developments of existing methods. Although several works partially alleviate this issue by using synthetic datasets or small-scale real datasets. The haze intensity distribution bias and scene homogeneity in existing datasets limit the generalization ability of these methods, particularly when encountering images with previously unseen haze intensities. 

In this work, we present LMHaze, a large-scale, high-quality real-world dataset. LMHaze comprises paired hazy and haze-free images captured in diverse indoor and outdoor environments, spanning multiple scenarios and haze intensities. It contains over 5K high-resolution image pairs, surpassing the size of the biggest existing real-world dehazing dataset by over 25 times.
Meanwhile, to better handle images with different haze intensities, we propose a mixture-of-experts model based on Mamba (MoE-Mamba) for dehazing, which dynamically adjusts the model parameters according to the haze intensity. Moreover, with our proposed dataset, we conduct a new large multimodal model (LMM)-based benchmark study to simulate human perception for evaluating dehazed images. Experiments demonstrate that LMHaze dataset improves the dehazing performance in real scenarios and our dehazing method provides better results compared to state-of-the-art methods. The dataset and code are available at our \href{https://github.com/wangzrk/LMHaze}{project page}.

\end{abstract}




\maketitle

\section{Introduction}

\begin{figure*}[ht]
    \centering
\vspace{-3mm}     \begin{tikzpicture}
     \node[anchor=south west,inner sep=0] (image) at (0,0) {\includegraphics[width=0.70\linewidth]{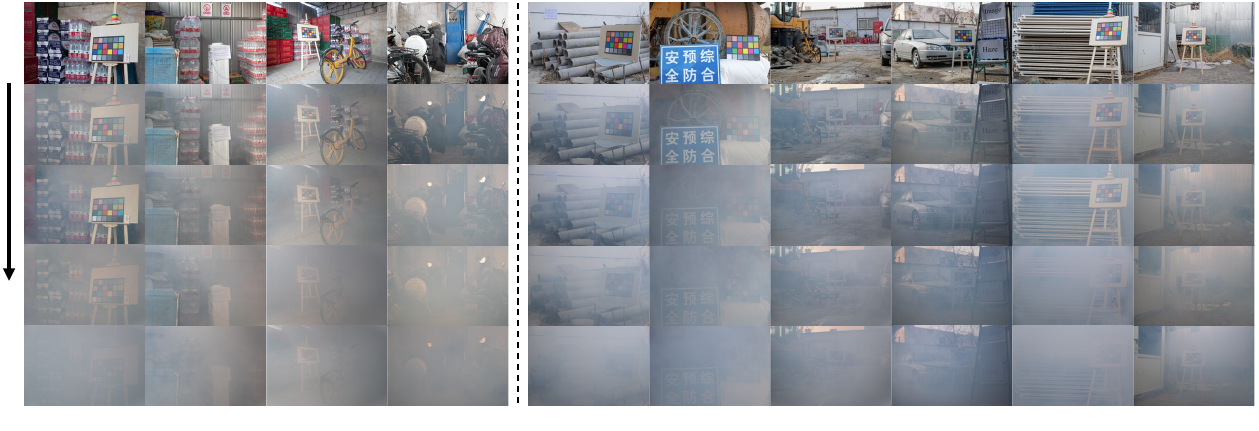}};
        \begin{scope}[x={(image.south east)},y={(image.north west)}]
            \draw (0.215, 0.02) node {(a) Indoor scenes};
            \draw (0.715, 0.02) node {(b) Outdoor scenes};
            \draw (-0.008, 0.22) node[rotate=90] {\small dense haze};
            \draw (-0.008, 0.90) node[rotate=90] {\small haze-free};
        \end{scope}
    \end{tikzpicture}
    \vspace{-15pt}
    \caption{\small \it 
    Examples of our LMHaze dataset. It includs 5,040 pairs of real-world hazy and haze-free images, spanning both indoor and outdoor scenarios. Each column illustrates a specific scenario, presenting the haze-free image alongside its associated hazy counterparts with varying haze intensities from top to bottom. 
    }
    \Description{Examples from the LMHaze dataset showing indoor and outdoor scenes with varying intensities of haze and their corresponding clear images. }
    \label{fig:showdataset}
    \vspace{-14pt}
\end{figure*}
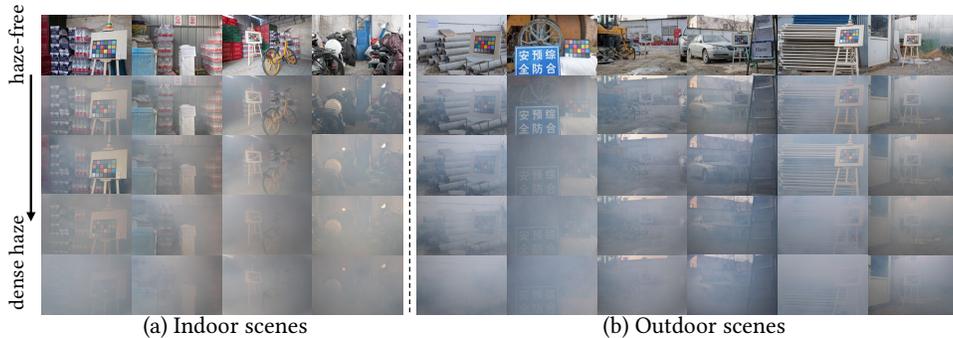

Image dehazing aims to recover clean images from hazy ones, where the improved images are essential for high-level tasks~\cite{downstream_understanding, downstream_detection, downstream_segmentation,liu2024lightweight,pan2019joint}.
%
In recent years, learning-based dehazing methods have achieved significant progress  \cite{AECRNet_CVPR21,image_dehaze_1,image_dehaze_2,image_dehaze_3,image_dehaze_5,image_dehaze_6,image_dehaze_7,image_dehaze_8,image_dehaze_9,DehazeNet_TIP16,FFANet_AAAI20, AODNet_ICCV17}. The hazy images are usually synthesized based on the atmospheric scattering model (ASM) \cite{atom_scatter_model}, which is expressed as,
\begin{align}
\vI(x) = \vJ(x)\vt(x) + \vA(1 - \vt(x)) \ ,
\end{align}
where $\vt(x) = e^{-\beta \vd(x)}$, and $\vI$, $\vJ$, $\vA$, $\vt$, and $\vd$ denote the hazy image, haze-free image, intensity of global atmospheric light, medium transmission map, and depth map, respectively. Here, $x$ is the pixel position and $\beta$ is the atmosphere scattering coefficient.

While the ASM allows the generation of large-scale hazy images from haze-free counterparts, it faces two limitations: (i) Obtaining dense and precise depth for high-quality hazy image synthezis, particularly in outdoor scenarios, remains a challenging task. (ii) It is struggle for ASM  to faithfully represent real-world haze scenes, particularly when dealing with complex lighting conditions~\cite{ASM_challenge,Dense-Haze,NH-Haze}. Therefore, models trained on synthetic datasets yield unsatisfactory results when applied to real-world scenarios, especially in cases involving dense haze~\cite{Dense-Haze} and non-homogeneous haze~\cite{NH-Haze}.

Recently, several real-world dehazing datasets~\cite{I-Haze,O-Haze,NH-Haze,Dense-Haze,BeDDE} have been introduced, along with benchmarks for training and evaluating real-world dehazing algorithms. However, certain limitations persist. Firstly, existing real-world datasets are relatively small in terms of both resolution and scale. Representative datasets such as I-Haze~\cite{I-Haze}, O-Haze~\cite{O-Haze}, and Dense-Haze~\cite{Dense-Haze} have no more than 50 training samples (see \cref{tab:dataset}), which potentially lead to model overfitting. Secondly, the dataset exhibits limited diversity in haze intensity, with similar levels of haze across images. As a result, models trained on such datasets suffer performance degradation when applied to images with varying degrees of haze. Moreover, semantic annotations are rarely provided by these datasets,  which poses an inconvenience when assessing the benefits of dehazing methods for downstream tasks like object detection, semantic segmentation, and image captioning. Therefore, collecting a large-scale, high-quality real-world dehazing dataset represents a challenging yet valuable endeavor for the research community.

In this paper, we first present a real-world dehazing dataset, LMHaze. Then, we propose a baseline method, MoE-Mamba, for benchmarking our dataset. The LMHaze has following advantages:
\textbf{i) Large Scale and High Resolution.} It contains 5,040 hazy/haze-free image pairs, while other real-world dehazing datasets such as I-Haze \cite{I-Haze}, O-Haze \cite{O-Haze}, Dense-Haze \cite{Dense-Haze}, and NH-Haze \cite{NH-Haze} contain only 35, 45, 33, and 55 image pairs, respectively. Moreover, our images with resolutions up to 5472×3648, surpass those of all current datasets.  
\textbf{ii) Multi-intensity.} 
It includes 8 haze intensities within each scene, providing a notable contrast to existing real-world datasets which often overlook the diversity of haze levels across images. 
\textbf{iii) Semantic Annotation.} 
It contains multi-type of hand-crafted annotations, including labels for object detection, semantic segmentation, and captioning. 

Based on this dataset, we conduct a comprehensive benchmark study of state-of-the-art (SOTA) dehazing methods. In addition to commonly used metrics such as PSNR and SSIM, we employ LMM-based evaluation metrics to simulate human visual perception. Moreover, we assess the performance of these methods in various downstream tasks involving dehazed images, thereby highlighting their practical applicability. Notably, our observations reveal that SOTA methods struggle to achieve satisfactory results when handling multi-intensity hazy images. As a result, we are motivated to develop a new dehazing baseline to address the challenges posed by varying levels of haze intensity.


In this paper, we present the MoE-Mamba framework for image dehazing. We achieve robust intensity-aware dehazing performance with three key components, \ie, i) an LMM-based intensity-aware block, ii) a Mixture-of-Experts (MoE) block, and iii) a State space model block. Specifically: 
i) By employing a training-free LMM-based intensity-aware module, we estimate haze intensity, without relying on additional intensity labels for supervision; 
ii) By using the MoE block based on the expert mechanism~\cite{MoE}, our model dynamically adapts network parameters with the estimated haze intensity to handle hazy images across diverse intensities; 
iii) To explore valuable non-local information while maintaining linear computational complexity, our Mamba block integrates a selective scanning mechanism of Mamba~\cite{mamba}, leading to improved restoration results and adaptability to complex scenes. 

Our contributions are summarized as follows:
\vspace{-1.5mm}
\begin{itemize}
\item We introduce, LMHaze, a large-scale real-world dehazing dataset comprising high-resolution images across varying haze intensities, as well as semantically annotated labels for downstream task evaluations.

\item We conduct an extensive benchmark study using SOTA dehazing methods. Our evaluation metrics include PSNR, SSIM, LPIPS, and LMM-based measures, in addition to assessments related to downstream vision tasks.

\item We propose a intensity-aware baseline, \ie, MoE-Mamba framework. Our extensive experiments demonstrate the competitive performance of our method in both qualitative and quantitative evaluations.
\end{itemize}
\vspace{-2mm}

\section{Related Work}

\begin{table*}[t]
\vspace{-2mm}
\centering
\caption{\small \it
Comparison of image dehazing datasets. ‘Real’ denotes whether hazy images are captured from real-world haze scenes. ‘I$\&$O’ denotes datasets that include both indoor and outdoor scenes. ‘Annotation’ signifies whether annotations for downstream tasks (\ie, detection, segmentation, and captioning) are provided. ‘Multi Inten’ indicates datasets that cover multiple intensities of haze. Lastly, ‘GT’ refers to datasets that include ground truth information for each scene.
}
\vspace{-10pt}
\setlength{\tabcolsep}{6pt}
\begin{tabular}{l|cccccccccc}
\Xhline{4\arrayrulewidth}
&&&&&&\multicolumn{3}{c}{Annotation}\\
\cline{7-9}
\multirow{-2}*{Dataset} & \multirow{-2}*{Year} & \multirow{-2}*{Scale} & \multirow{-2}*{Resolution} & \multirow{-2}*{Real} 
& \multirow{-2}*{I$\&$O} & Detection & Segmentation & Caption 
& \multirow{-2}*{Multi Inten.} & \multirow{-2}*{GT}\\
\Xhline{4\arrayrulewidth}
FRIDA \cite{FRIDA} & 2010 & 90 & 640$\times$480 & \ding{55} & O & \ding{55} & \ding{55} & \ding{55} & \ding{51} & \ding{51}\\
FRIDA2 \cite{FRIDA} & 2012 & 330 & 640$\times$480 & \ding{55} & O & \ding{55} & \ding{55} & \ding{55} & \ding{51} & \ding{51}\\
D-HAZY \cite{D-hazy} & 2016 & 1400 & 1K--2K & \ding{55} & I & \ding{55} & \ding{55} & \ding{55} & \ding{55} & \ding{51}\\
HazeRD \cite{HazeRD} & 2017 & 15 & 2K--4K & \ding{55} & O & \ding{55} & \ding{55} & \ding{55} & \ding{51} & \ding{51}\\
ITS \cite{RESIDE} & 2018 & 13990 & 640$\times$480 & \ding{55} & I & \ding{55} & \ding{55} & \ding{55} & \ding{51} & \ding{51}\\
SOTS \cite{RESIDE} & 2018 & 500 & 640$\times$480 & \ding{55} & I & \ding{55} & \ding{55} & \ding{55} & \ding{51} & \ding{51}\\
OTS \cite{RESIDE} & 2018  & 72135 & 640$\times$480 & \ding{55} & O & \ding{55} & \ding{55} & \ding{55} & \ding{51} & \ding{51}\\
haze4K \cite{haze4K} & 2021 & 4000 & 400$\times$400 & \ding{55} & I\&O & \ding{55} & \ding{55} & \ding{55} & \ding{51} & \ding{51}\\
\hline
HSTS \cite{RESIDE} & 2018 & 20 & 640$\times$480 & \ding{51} & O & \ding{55} & \ding{55} & \ding{55} & \ding{55} & \ding{55}\\
RTTS \cite{RESIDE} & 2018 & 4322 & 640$\times$480 & \ding{51} & O & \ding{51} & \ding{55} & \ding{55} & \ding{55} & \ding{55}\\
O-HAZE \cite{O-Haze} & 2018 & 45 & 2K--4K & \ding{51} & O & \ding{55} & \ding{55} & \ding{55} & \ding{55} & \ding{51}\\
I-HAZE \cite{I-Haze} & 2018  & 35 & 2K--4K & \ding{51} & O & \ding{55} & \ding{55} & \ding{55} & \ding{55} & \ding{51}\\
Dense-HAZE \cite{Dense-Haze} & 2019 & 55 & 2K--4K & \ding{51} & I\&O & \ding{55} & \ding{55} & \ding{55} & \ding{55} & \ding{51}\\
NH-HAZE \cite{NH-Haze} & 2020 & 55 & 2K--4K & \ding{51} & O & \ding{55} & \ding{55} & \ding{55} & \ding{55} & \ding{51}\\
BeDDE \cite{BeDDE} & 2020 & 208 & 1K-2K & \ding{51} & O & \ding{55} & \ding{55} & \ding{55} & \ding{55} & \ding{51}\\
\Xhline{4\arrayrulewidth}
\textbf{Ours} & \textbf{2024} & \textbf{5,040} & \textbf{5472$\times$3648} & \ding{51} & \textbf{I\&O} & \ding{51} & \ding{51} & \ding{51} & \ding{51} & \ding{51}\\
\Xhline{4\arrayrulewidth}
\end{tabular}
\vspace{-10pt}
\label{tab:dataset}
\end{table*}


\noindent{\textbf{Image Dehazing Dataset.}}
Learning-based dehazing methods require paired hazy/haze-free images, but collecting such data in real-world scenarios is challenging. Early works synthesize haze using datasets with depth information \cite{Middleblur, NYU-Depth, Cityscapes} based on the ASM model. Despite achieving satisfactory results, the domain gap between synthetic and real hazy images limits model performance.
To address this, researchers have assembled real-world dehazing datasets. One approach simulates hazy scenes using haze generators~\cite{I-Haze, O-Haze, NH-Haze, Dense-Haze}. However, these datasets are small and lack varying haze intensities. Another approach uses fixed-view cameras to capture hazy and haze-free scenes, like BeDDE \cite{BeDDE}, which collected 208 images over a year. Due to the challenges of outdoor conditions, dataset sizes remain limited. Thus, we introduce a large-scale, real-world dataset to tackle this issue in dehazing research.

\noindent{\textbf{Mixture-of-Experts Model.}}
The Mixture-of-Experts (MoE) model combines multiple expert models \cite{MoE}, each specialized in specific tasks. A gating mechanism dynamically selects their outputs to improve performance and generalization. Within fixed computational costs, increasing experts enhances model capacity, widely validated in computer vision \cite{moe_cv_0,moe_cv_1,moe_cv_2,moe_cv_3,Yang_2024_CVPR}. We use MoE model to adjust model parameters dynamically under varying haze intensities, improving robustness.

\noindent{\textbf{State Space Model.}}
State Space Models (SSMs) \cite{state_space_model} show great potential for modeling long-range dependencies in NLP \cite{state_space_model_NLP_0,state_space_model_NLP_1,state_space_model_NLP_2}. Mamba \cite{mamba}, an improved SSM with a selective scanning mechanism (S6), remembers relevant information while maintaining linear complexity. Mamba has been applied to vision tasks like object detection~\cite{state_space_model_detection_0}, image classification~\cite{state_space_model_imageclassification_0}, and semantic segmentation~\cite{state_space_model_segmentation_0}. We explores its potential in image dehazing as a baseline for future research.

\vspace{-2mm}
\section{LMHaze Dataset}

We introduce the image capture settings in Sec.~\ref{sec:setup}, and post-processing and labeling  of our LMHaze dataset in Sec.~\ref{sec:post}.

\subsection{Image capture settings}

\noindent{\textbf{Capture Setup.}}
\label{sec:setup}
Our image capture setup is as follows. We capture static scenes (both indoor and outdoor) to avoid misalignment due to scene motion. We use a Canon EOS-1D X Mark II camera equipped with a Canon EF 24-70mm f2.8L II USM lens. To minimize camera shake, the camera is mounted on a sturdy tripod and a wireless shutter trigger is used to control image acquisition. RAW images are acquired to avoid variables introduced by ISP processing and to ensure consistency in the captured images. Similar to the haze generation strategy in \cite{O-Haze}, we use a professional haze generator, which rapidly passes smoke oil through a high-temperature heated tube to form a white gaseous smoke spray.

\begin{figure*}[ht]
    \centering
 \vspace{-2mm}    \begin{tikzpicture}
    \node[anchor=south west,inner sep=0] (image) at (0,0) {\includegraphics[width=0.8\linewidth]{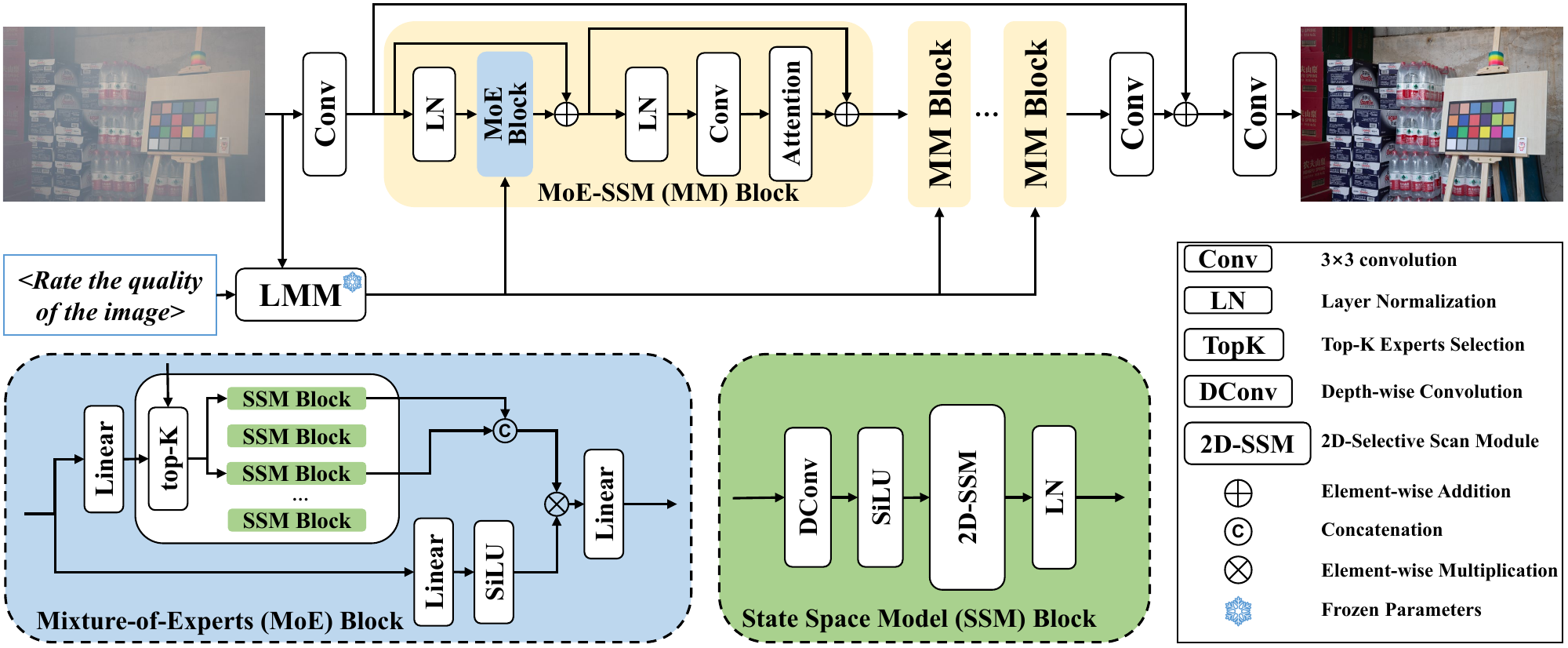}};
        \begin{scope}[x={(image.south east)},y={(image.north west)}]
            \draw (0.087, 0.655) node {$\mbox{\bf I}_{\text{h}}$};
            \draw (0.917, 0.655) node {$\mbox{\bf I}_{\text{dh}}$};
            \draw (0.253, 0.58) node {$\hat{\mbox{\bf p}}$};
            \draw (0.235, 0.79) node {\small $\mbox{\bf F}_\text{s}$};
            \draw (0.565, 0.79) node {\small $\mbox{\bf F}_\text{s}^{1}$};
            \draw (0.695, 0.79) node {\small $\hat{\mbox{\bf F}}_\text{s}$};
            \draw (0.775, 0.785) node {\small $\mbox{\bf F}_\text{a}$};
            \draw (0.115, 0.40) node {\small $\hat{\mbox{\bf p}}$};
            \draw (0.017, 0.17) node {\small $\mbox{\bf F}_\text{ln}$};
            \draw (0.42, 0.18) node {\small $\mbox{\bf F}_\text{d}$};
        \end{scope}
    \end{tikzpicture}
    \vspace{-12pt}
    \caption{\small \it Overview of our MoE-Mamba framework. Given a hazy image $\mbox{\bf I}_{\text{h}}$, our goal is to restore a clean image $\mbox{\bf I}_{\text{dh}}$. We first feed prompt and $\mbox{\bf I}_{\text{h}}$ to the large multimodal model (\text{LMM})~\cite{Q-Align}, and outputting the image degradation prior $\hat{\mbox{\bf p}}$. Meanwile, features $\mbox{\bf F}_\mathrm{s}$ are extracted by a 3$\times$3 conv layer. 
    Then, the $\hat{\mbox{\bf p}}$ and $\mbox{\bf F}_\mathrm{s}$ are passed to multiple MoE-SSM (MM) Blocks to get refined feature $\hat{\mbox{\bf F}}_\mathrm{s}$. Then, element-wise addition is used to get $\mbox{\bf F}_\mathrm{a}$.
    The $\mbox{\bf I}_{\text{dh}}$ are reconstructed based on features $\mbox{\bf F}_\mathrm{a}$. The key elements of our MM Block, \ie, the Mixture-of-Experts (MoE) Block and the State Space Model (SSM) Block, are shown in the \textcolor{lightBlue}{blue} and the \textcolor{lightGreen}{green} dotted box.
    }
    \label{fig:framework}
    \Description{Overview of our MoE-Mamba framework.}
    \vspace{-15pt}
\end{figure*}

\noindent{\textbf{Capture Process.}} 
With the capture setup, we collect hazy images of various scenes, both indoor and outdoor. For each scene, before generating the haze, we first capture a haze-free image which we use as the ground truth. Then, we start generating haze and use a fan to spread the haze into the scene. To capture haze with multiple haze intensities, we capture a series of sequential images with a time interval of \textbf{0.5-5} minutes between sequential images. The haze generator is periodically turned on and off during the entire image sequence acquisition. We capture all intermediate states where the haze gradually changed from none to dense, and these intermediate state images reflect different intensities of haze levels. To ensure alignment between hazy and haze-free images, the entire scene remains static during capture. For each scene, the camera is static and all camera settings (e.g., ISO, exposure, focus, white balance, exposure compensation) are fixed. We take a haze-free image again at the end of each scene (after the haze has completely disappeared) as an additional reference image.

More details of camera's optical parameters and correction operations are in supplementary materials. \cref{fig:showdataset} shows examples of outdoor and indoor hazy images, where each example contains a series of images with varying haze intensities. Here, 8,326 hazy/haze-free pairs are collected. Post-processing is required for the collect raw data to building a high-quality dataset.

\subsection{Image post-processing}
\label{sec:post}
With the RAW data, we first remove unwanted images manually, such as images with negligible haze. Then, we verify the alignment of hazy and haze-free images. We employ a widely used disparity estimation method \cite{disparity} to measure the shift between the hazy  and haze-free image, as well as the reference image. Consequently, we exclude scenes containing images with an average disparity exceeding 1 pixel, resulting in a remaining set of 5,040 image pairs.
%
%
Next, we use Adobe Photoshop Lightroom Classic to correct the data for white balance, color correction, gamma correction, etc. We maintain the consistency of the haze color in all scenes. All images are then converted to a lossless PNG format with a spatial resolution of 5472×3648 with a 24-bit color depth. 

To demonstrate that collected images have diverse haze intensity levels, we borrow the power of existing large multimodal models (\ie~Q-Align \cite{Q-Align}) to eliminate subjective factors. 
By using the quality scores~\cite{Q-Align} as the criteria for intensity classification, we calculate quality scores of 5,040 images with the prompt \textit{\tt \{Rate the quality of the image.\}}. Our dataset can be split into three subsets with different haze intensity levels, where 920 is light haze (18.3\%), 3,182 is medium haze (63.1\%), and 938 is dense haze (18.6\%).

\noindent{\textbf{Image Annotation.}} 
To further evaluate the performance of dehazing methods on downstream tasks, we provide three types of annotations, \ie, segmentation, detection, and captioning. To improve annotation efficiency, we first apply a SOTA segmentation, object detection model~\cite{SAM,DEQDet} to each haze-free image in our dataset, which provides roughly accurate masks and bounding boxes for most common objects. We then manually modify incorrect labels and add annotations for undetected objects. For image captions, we feed haze-free images directly into the LMM model to obtain them and manually fine-tune captions with incorrect semantics.

\section{Proposed MoE-Mamba}
In this section, we propose the overall network architecture (Sec.~\ref{sec:ov}), main modules and loss functions (Sec.~\ref{sec:prior}) of our MoE-Mamba.


\subsection{Overview}
\label{sec:ov}

Given a degraded image $\mathbf{I}_{\text{h}}$ affected by haze, our goal is to restore the clean image $\mathbf{I}_{\text{dh}}$. As shown in \cref{fig:framework}, our method adaptively restores the dehazed image $\mathbf{I}_{\text{dh}}$ with the help of image degradation prior $\mathbf{\hat{p}}$ obtained from an intensity-aware block based on a pre-trained large multimodal model (LMM) $\mathbf{M}(\cdot , \cdot )$. To obtain $\mathbf{\hat{p}}$ containing a representation of the severity of image degradation, we first set up a text prompt $\mathbf{Q}$, and then input hazy image $\mathbf{I}_{\text{h}}$ and $\mathbf{Q}$ into $\mathbf{M}(\cdot , \cdot )$ to query the image probability distribution $\mathbf{p}$. We then convert $\mathbf{p}$ into specific rating levels, obtaining the image degradation prior $\mathbf{\hat{p}}$.

We restore $\mathbf{I}_{\text{h}}$ to $\mathbf{I}_{\text{dh}}$ through the following steps.
Given $\mathbf{I}_{\text{h}}$, we first utilize a 3$\times$3 convolution to obtain shallow feature $\mathbf{F}_{\text{s}}$.
With $\mathbf{F}_{\text{s}}$ and $\mathbf{\hat{p}}$, we employ multiple MoE-SSM (MM) blocks to obtain the feature $\hat{\mathbf{F}}_{\text{s}}$. 
Then, element-wise addition is used to combine the features and get $\mathbf{F}_{\text{a}}$.
Next, a dehazed image $\mathbf{I}_{\text{dh}}$ can be reconstructed from $\mathbf{F}_{\text{a}}$ with a 3$\times$3 convolution projection layer.

\begin{table*}[t]
\vspace{-2mm}
    \caption{\small \it Comparisons of SOTA dehazing methods on LMHaze. We highlight the best method in \textbf{bold}.}
    \vspace{-10pt}
    \label{tab:benchmark}
    \centering
    \setlength{\tabcolsep}{3pt}
        \begin{tabular}{l |c |c c c c|c c c c| c c c c}
            \Xhline{4\arrayrulewidth} 
            &
            &\multicolumn{4}{c|}{LMHaze}
            &\multicolumn{4}{c|}{LMHaze-outdoor}
            &\multicolumn{4}{c}{LMHaze-indoor}\\
            \multirow{-2}*{Method}
            &\multirow{-2}*{Venue}
            & PSNR~$\textcolor{black}{\uparrow}$ & SSIM~$\textcolor{black}{\uparrow}$ & QAS$\textcolor{black}{\uparrow}$ & LPIPS~$\textcolor{black}{\downarrow}$
            & PSNR~$\textcolor{black}{\uparrow}$ & SSIM~$\textcolor{black}{\uparrow}$ & QAS$\textcolor{black}{\uparrow}$ & LPIPS~$\textcolor{black}{\downarrow}$			
            & PSNR~$\textcolor{black}{\uparrow}$ & SSIM~$\textcolor{black}{\uparrow}$ & QAS$\textcolor{black}{\uparrow}$ & LPIPS~$\textcolor{black}{\downarrow}$ \\
            \Xhline{4\arrayrulewidth} 
            DCP~\cite{DCP_TPAMI10}              & TPAMI'10   & 9.75  & 0.4752 & 2.895 & 0.4530 & 9.52 & 0.4790  & 2.776 & 0.4710 & 10.10 & 0.4689  & 3.066 & 0.4268\\ 
            DehazeNet~\cite{DehazeNet_TIP16}    & TIP'16     & 12.76 & 0.6130 & 2.764 & 0.4474 & 12.97 & 0.6294 & 2.671 & 0.4668 &  12.45 & 0.5890 & 2.900 & 0.4191\\ 
            AODNet~\cite{AODNet_ICCV17}         & ICCV'17    & 14.65 & 0.5797 & 2.963 & 0.4039 & 15.07 & 0.6587 & 2.861 & 0.3974 & 14.59 & 0.5536 & 3.173 & 0.4085\\ 
            GridDehazeNet~\cite{GridDehazeNet_ICCV19}   & ICCV'19  & 15.91 & 0.6785 & 3.248 & 0.3486 & 15.70 & 0.6646 & 3.167 & 0.3679 & 16.16 & 0.6812 & 3.315 & 0.3412\\ 
            FFANet~\cite{FFANet_AAAI20}         & AAAI'20    & 16.34 & 0.7032 & 3.326 & 0.2643 & 16.46 & 0.6766 & 3.297& 0.2727 & 16.13 & 0.7156 & 3.367 & 0.2521\\ 
            AECRNet~\cite{AECRNet_CVPR21}       & CVPR'21    & 15.80 & 0.4660 & 3.472 & 0.3265 & 17.39 & 0.5897 & 3.337 & 0.2970 & 14.28 & 0.4074 & 3.509 & 0.3482\\ 
            Dehamer~\cite{Dehamer}              & CVPR'22    & 15.76 & 0.5780 & 3.523 & 0.3196 & 16.00 & 0.5786 & 3.550 & 0.3250 & 15.51 & 0.5696 & 3.486 & 0.3116\\ 
            DehazeFormer~\cite{DehazeFormer}    & TIP'23     & 17.70 & 0.7628 & 3.588 & 0.2215 & 18.04 & 0.7570 & 3.578 & 0.2161 & 17.67 & 0.7690 & 3.604 & 0.2299\\
            SLP~\cite{image_dehaze_8}                       & TIP'23     & 10.00 & 0.5011 & 2.940 & 0.4507 & 9.61 & 0.5017 & 2.734 & 0.4692 & 10.56 & 0.5002 & 3.092 & 0.4198\\
            MB-TaylorFormer~\cite{MBTaylorFormer}& ICCV'23    & 17.73 & 0.7286 & 3.545 & 0.2253 & 17.75 & 0.7430 & 3.487 & 0.2274 & 17.35 & 0.7105 & 3.629 & 0.2223\\
            DEANet~\cite{DEANet}                 & TIP'24     & 17.96 & 0.7453 & 3.281 & 0.2686 & 18.03 & 0.7521 & 3.295 & 0.2694 & 17.92 & 0.7296 & 3.263 & 0.2675\\
            LDR~\cite{LDR_CVPR24}                  & CVPR'24    & 17.97 & 0.7457 & 3.586 & 0.2107 & 17.95 & 0.7548 & 3.504 & 0.2119 & 17.98 & 0.7323 & 3.706 & 0.2072\\    
            MambaIR~\cite{MambaIR}               & ECCV'24          & 17.94 & 0.7364 & 3.509 & 0.2540 & 17.93 & 0.7509 & 3.445 & 0.2549 & 17.96 & 0.7554 & 3.603 & 0.2532\\
            \hline
            \textbf{Ours} & - & \textbf{18.49} & \textbf{0.7860} & \textbf{3.862} & \textbf{0.1928} & \textbf{18.56} & \textbf{0.7909} & \textbf{3.672} & \textbf{0.1953} & \textbf{18.48} & \textbf{0.7744} & \textbf{3.955} & \textbf{0.1891}\\
            \Xhline{4\arrayrulewidth}
    \end{tabular}		
\end{table*}

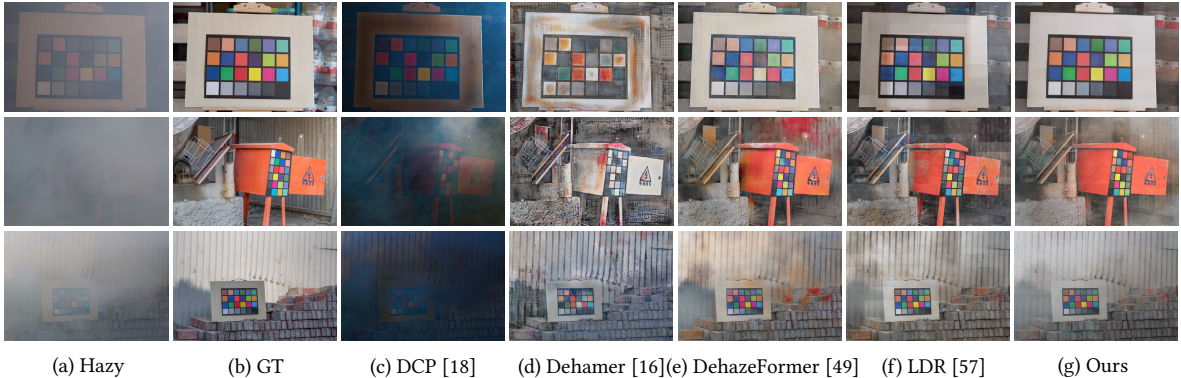
\begin{figure*}[t]
\vspace{-2mm}
    \centering
    \begin{tikzpicture}
    \node[anchor=south west,inner sep=0] (image) at (0,0) {\includegraphics[width=0.88\linewidth]{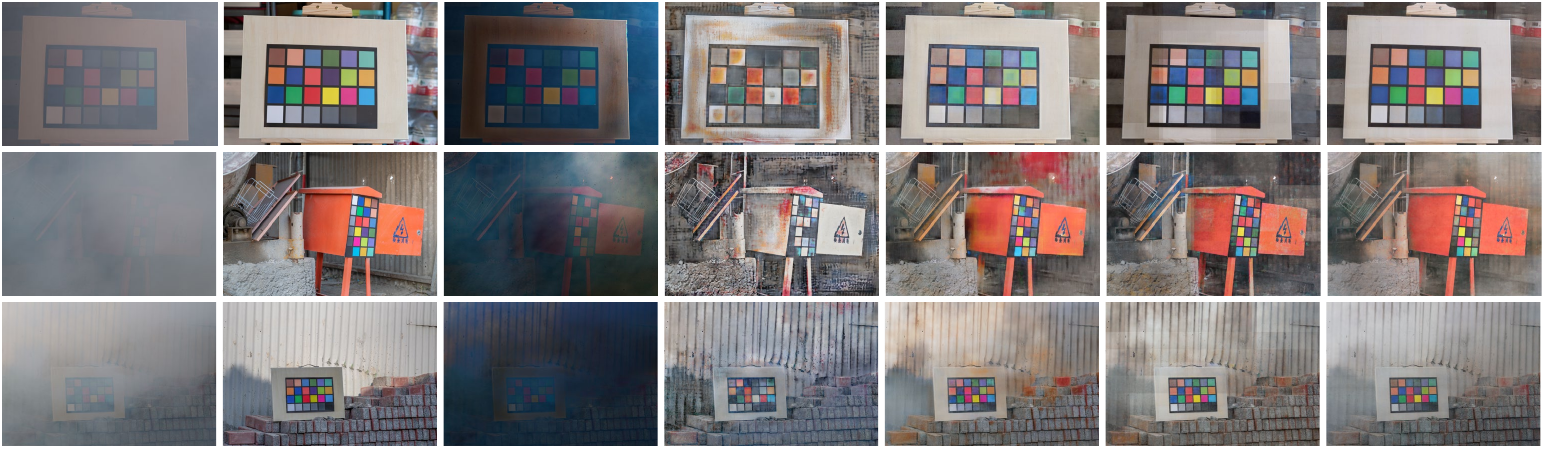}};
        \begin{scope}[x={(image.south east)},y={(image.north west)}]
            \draw (0.073,  -0.07) node {\small (a) Hazy};
            \draw (0.215,  -0.07) node {\small (b) GT};
            \draw (0.357,  -0.07) node {\small (c) DCP~\cite{DCP_TPAMI10}};
            \draw (0.50,  -0.07) node {\small (d) Dehamer~\cite{Dehamer}};
            \draw (0.645,  -0.07) node {\small (e) DehazeFormer~\cite{DehazeFormer}};
            \draw (0.79,  -0.07) node {\small (f) LDR~\cite{LDR_CVPR24}};
            \draw (0.925,  -0.07) node {\small (g) Ours};
        \end{scope}
    \end{tikzpicture}
    \vspace{-15pt}
    \caption{\small \it Example results on LMHaze dataset. (a)-(b) are the hazy image and corresponding haze-free image. (c)-(g) show the dehazing results of different SOTA dehazing models. Our model achieves the best dehazing performance while restoring the most accurate color details.}
    \Description{Example results on LMHaze dataset.}
    \vspace{-13pt}
    \label{fig:benchmark}
\end{figure*}

\subsection{MoE-Mamba}
\label{sec:prior}
\noindent{\textbf{LMM-based Intensity-aware Block.}}
We leverage the powerful low-level visual perception capabilities of the LMM (\eg,~Q-Align \cite{Q-Align}) by designing a prompt $\mathbf{Q}$ to assess the quality of hazy image. We follow the specifically designed for low-quality images in \cite{Q-Align} as $\mathbf{Q}$ = \textit{{{\tt <Rate the quality of the image>}}}.
The outputting can be denoted as,
\begin{equation}
    \mathbf{p} = \mathbf{M}(\mathbf{I}_{\text{h}} \ , \mathbf{Q}) \ , \quad  \mathbf{p} \in \mathbb{R}^{L} \ ,
\end{equation}
where $\mathbf{M}(\cdot, \cdot)$ is the LMM, $L$ is the length of all tokens in the LMM, and $\mathbf{p}$ is the outputting probability distribution. 

Following~\cite{Q-Align}, we link $\mathbf{p}$ with certain rating levels.
First, we build a candidate level set $\cG$. Specifically, we choose \textit{\tt \{bad, poor, fair, good, excellent\}} as standard rating levels, which is defined by ITU~\cite{itu}. Besides, to fit the dehazing task, we also include additional reference rating levels that describe haze intensities (\eg,~\textit{\tt clear, fog, mist}). The built level set $\cG$ contains $N$ rating levels. 
%
Then, we define a mapping $G(\cdot)$ from the text-defined rating levels to the index of the probability, \ie, $G: \cG_i \rightarrow j$, where $i \in [1, N]$ and $j \in [1, L]$. A close-set softmax is conducted on $\mathbf{p}$ to get the degradation prior, 

\begin{equation}
    \mathbf{\mathbf{\hat{p}}}(i) = \mathrm{Softmax}\Big(\mathbf{p}\big(G(\cG_i)\big)\Big) \ ,
\end{equation}
where $\mathbf{\mathbf{\hat{p}}}(i)$ is probability of the $i$-th rating level. 

\noindent{\textbf{MoE-SSM (MM) Block.}} The degradation prior is then feed to the MM block for
dynamically changing the model parameters and achieving intensity-aware image dehazing:
\begin{align}
    \hat{\mathbf{F}}_{\text{s}} &= \text{MM}(\mathbf{F}_\text{s},\mathbf{\hat{p}}, H) + \delta \cdot \mathbf{F}_s \ ,
\end{align}
where $\mathbf{F}_\text{s}$ is the feature obtained by convolution layer from $\mathbf{I}_h$, $\delta$ is a learnable scale factor, and $H$ is the number of the employed MM blocks. Here, $\text{MM}(\cdot ,\cdot , \cdot )$ denotes the processing of muiltiple MM blocks as shown in \cref{fig:framework}. After passing $H$ MM blocks, we obtain the refined feature $\hat{\mathbf{F}}_{\text{s}}$. Taking a single MM block as an example, it contains two key elements, \ie, the MoE block and the SSM block. 

\noindent{\textbf{Mixture-of-Experts (MoE) Block.}} 
Given $\mathbf{F}_\text{s}$ , we first pass them through a LayerNorm, following~\cite{state_space_model_imageclassification_0}, to extract long-range dependencies, and resulting in $\mathbf{F}_\text{ln}$. Based on the top-$K$ experts selection mechanism, we devise the MoE block to dynamically select parameters for extracting degradation feature $\mathbf{F}_\text{d}$:
\begin{align}
    \mathbf{F}_{\text{d}} = \text{Linear}\Big(\text{TK}\big(\mathbf{\hat{F}}_{\text{ln}}, \mathbf{\hat{p}}\big) \otimes \text{SiLU}\big(\text{Linear}(\mathbf{F}_\text{ln})\big)\Big)  \ ,
\end{align}
where $\text{TK}(\cdot , \cdot )$ represents the top-$K$ experts selection mechanism, and $\mathbf{\hat{F}}_{\text{ln}} =\text{Linear}( \mathbf{F}_{\mathrm{ln}})$. Here, $\text{Linear}(\cdot)$ and $\otimes$ are fully connected layer and element-wise multiplication respectively.
Following traditional Transformer architecture \cite{transformer_architecture}, we add $\mathbf{F}_\text{s}$ to $\mathbf{F}_\text{d}$, and pass the result through a LayerNorm, a conv layer, and an attention layer sequentially, to obtain the output, \eg, $\mathbf{F}_\text{s}^{1}$ for the $1$-st MM block .

In the process of selecting the top-$K$ experts using the function $\text{TK}(\cdot, \cdot)$, we utilize multiple SSM blocks (SSBs). Considering the varying haze intensity levels across images, we adopt an adaptive strategy to choose different SSBs based on the degradation prior  $\mathbf{\hat{p}}$. Initially, we define $N$ candidate SSBs and dynamically select $K$ SSBs corresponding to the highest probabilities in $\mathbf{\hat{p}}$. These selected SSBs are then used to generate features specific to the haze condition. The selection process is defined as,
\begin{align}
\text{TK}(\mathbf{\hat{F}}_{\text{ln}}, \mathbf{\hat{p}}) &= \sum_{i=1}^{K} \text{SSB}_{v(i)}(\hat{\mathbf{F}}_{\text{ln}})\otimes \mathbf{\hat{p}}_{v(i)}, \ \ \ 
v = \arg\max_{K}(\mathbf{\hat{p}}) \ ,
\end{align}
where $v$ denotes the selected indices of the top-$K$ largest elements in $\mathbf{\hat{p}}$, $\mathbf{\hat{p}}_{v(i)}$ is $i$-th largest element in $\mathbf{\hat{p}}$, $\text{SSB}_{v(i)}(\cdot )$ is the $v(i)$-th SSB of all condidate SSBs, and $\otimes$ is element-wise multiplication.

\noindent{\textbf{State Space Model (SSM) Block.}}  
Inspired by~\cite{MambaIR}, we use the SSM block to model long-range pixel dependencies. Each SSM block sequentially includes a depthwise convolution layer $\text{DConv}(\cdot )$, a SiLU activation function~\cite{SiLU} $\text{SiLU}(\cdot )$, a 2D-SSM layer~\cite{MambaIR} $\text{2D-SSM}(\cdot )$, and a LayerNorm $\text{LN}(\cdot )$. Given the input feature $\hat{\mathbf{F}}_{\text{ln}}$, the SSM block $\text{SSB}(\cdot )$ can be formulated as: 
\begin{equation}
    \text{SSB}(\hat{\mathbf{F}}_{\text{ln}}) = \text{{LN}\Big(
    2D-SSM\big(
    SiLU(
    DConv}(\hat{\mathbf{F}}_{\text{ln}})
    )
    \big)
    \Big).
\end{equation}

\noindent{\textbf{Loss Function.}}
We adopt the Charbonnier loss \cite{char_loss} to optimize our network: 
\begin{equation}
\mathcal{L}= \sqrt{\lVert \mathbf{I}_{\text{c}} - {\mathbf{I}}_{\text{dh}} \rVert^{2}+ \varepsilon^{2}} \ , 
\end{equation}
where $\mathbf{I}_{\text{c}}$ and $\mathbf{I}_{\text{dh}}$ denote the ground truth and dehazed image, respectively. Here, $\varepsilon$ is a constant that is empirically set to $1 \times 10^{-3}$.

\section{Experiments and Analysis}

\subsection{Experiments settings}

\noindent{\textbf{Implementation Details.}} 
Our method is fully implemented using Pytorch, and all experiments are conducted on a single NVIDIA RTX 3090 GPU. Our MoE-Mamba employs a four-level encoder-decoder structure. From the first to the fourth level, the number of MM blocks is \{4,6,6,8\}, and the values of $N$ and $K$ are \{14,1,1,14\} and  \{7,1,1,7\}, respectively. Our MoE-Mamba is trained with batch size $4$ for $2 \times 10^{5}$ iterations using the AdamW optimizer. The learning rate is decayed from $2 \times 10^{-4}$ to $1 \times 10^{-5}$ following a cosine annealing strategy. During training, we randomly crop the images to a size of $256 \times 256$. No cropping is performed during testing. For LMM in our model, We use the Q-Align \cite{Q-Align}, which is fine-tuned based on mPLUG-Owl2 \cite{ye2023mplug}. Our dataset and code will be made available online for future studies and comparisons.

\noindent{\textbf{Dataset.}} We conduct experiments on the LMHaze and Dense-Haze~\cite{Dense-Haze} dataset. Our LMHaze contains 5,040 hazy/haze-free image pairs with different haze intensities, where 3,925 pairs are for training and 1,115 pairs are for testing. The division is performed at the pair level, ensuring that no scenes present in the training set are included in the test set. Based on the scenario types, the LMHaze can also be divided into 3,545 outdoor images and 1,495 indoor images. 
Dense-Haze~\cite{Dense-Haze} is a commonly used real-world dehazing dataset, which contains 55 hazy/haze-free image pairs. Following \cite{MBTaylorFormer}, We split 50 pairs for training and 5 pairs for testing.

\noindent{\textbf{Evaluation Metrics.}}
We use peak signal-to-noise ratio (PSNR), structural similarity (SSIM) and learned perceptual image patch similarity (LPIPS) as our evaluation metrics. Additionally, we evaluate dehazed images using LMM and choose their quality assessment scores~\cite{Q-Align} (QAS) as a supplement to our evaluation metrics. 

\noindent{\textbf{Baseline Methods.}}
We compare with 13 state-of-the-art (SOTA) dehazing methods, \ie, DCP \cite{DCP_TPAMI10}, DehazeNet \cite{DehazeNet_TIP16}, AODNet \cite{AODNet_ICCV17}, GridDehazeNet \cite{GridDehazeNet_ICCV19}, FFANet \cite{FFANet_AAAI20}, AECRNet \cite{AECRNet_CVPR21}, Dehamer \cite{Dehamer}, DehazeFormer \cite{DehazeFormer}, SLP \cite{image_dehaze_8}, MB-TaylorFormer \cite{MBTaylorFormer}, DEANet \cite{DEANet} and MambaIR \cite{MambaIR}, as well as the all-in-one method LDR \cite{LDR_CVPR24}. Except for DCP \cite{DCP_TPAMI10} and SLP \cite{image_dehaze_8}, all other deep learning-based methods are retrained on the LMHaze dataset. 

\subsection{Experimental results}

\noindent{\textbf{Comparison with SOTA methods.}}
The comparisons on the LMHaze dataset are shown in \cref{tab:benchmark} and \cref{fig:benchmark}. 
Compared to the second-best method LDR~\cite{LDR_CVPR24}, 
we have an increase of 2.9\%/6.7\% in the metric of PSNR/SSIM. Compared to the SOTA Mamba architecture \cite{MambaIR}, 
we obtain an improvement of 3.1\%/6.7\% in PSNR/SSIM. The fact indicates the benefit of our model that uses image degradation prior for dynamic parameter selection. Furthermore, we achieve the best result for the LMM-based evaluation metric QAS, which indicates that the dehazed images processed by our method with better visual perception. The comparison on the Dense-Haze dataset is shown in the right column (w/o LMHaze) of \cref{tab:mixdata}. For space limitation, we compared with the top-ranking methods in \cref{tab:benchmark}. Our method achieves the best performance.

\noindent{\textbf{Quality of Our LMHaze Dataset.}} 
To showcase the utility of our LMHaze dataset to the research community, we first combine LMHaze with a commonly used real-world dehazing dataset, Dense-Haze~\cite{Dense-Haze}. \cref{tab:mixdata} reports the performance of SOTA methods on the Dense-Haze testing set with and without using additional images from LMHaze for training. Compared to models trained only on Dense-Haze, adding LMHaze as additional training data significantly improves the performance of each SOTA method. 
For example, MambaIR~\cite{MambaIR} demonstrates a substantial improvement in PSNR, up to 0.62 dB. 
The LMHaze dataset exhibits high image quality, making it a valuable complement to real-world dehazing datasets. 
To further verify the image quality of the LMHaze dataset, we employ the LMM \cite{Q-Align} to assess image quality on both LMHaze and existing real-world dehazing datasets. This comparison and additional results can be found in the supplementary materials.

\subsection{Ablation studies and discussions}

\begin{table}[t]
\vspace{-2mm}
    \caption{\small \it We compare the top-ranking dehazing methods in \cref{tab:benchmark} on the Dense-Haze dataset, considering with and without using LMHaze as additional training data. Note, models trained with LMHaze achieve superior results. 
    }
    \setlength{\tabcolsep}{5pt}
    \vspace{-10pt}
    \label{tab:mixdata}
    \centering
        \begin{tabular}{l |c c | c c}
            \Xhline{4\arrayrulewidth} 
             &\multicolumn{2}{c|}{w LMHaze}
            &\multicolumn{2}{c}{w/o LMHaze} \\
            \multirow{-2}*{Dense-Haze dataset}
            & PSNR~$\textcolor{black}{\uparrow}$ & SSIM~$\textcolor{black}{\uparrow}$ 
            & PSNR~$\textcolor{black}{\uparrow}$ & SSIM~$\textcolor{black}{\uparrow}$ \\
            \Xhline{4\arrayrulewidth}
            DEANet~\cite{DEANet} & {15.47}  & {0.4702} & 15.22 & 0.4078 \\
            LDR~\cite{LDR_CVPR24} & 15.79  & 0.5342 & 15.39 & 0.4477 \\            
            MambaIR~\cite{MambaIR} & {15.76} & {0.4724} & 15.14 & 0.4518 \\
            \textbf{Ours} & \textbf{15.81} & \textbf{0.5375} & \textbf{15.57} & \textbf{0.4801} \\ 
            \Xhline{4\arrayrulewidth}
    \end{tabular}		
    \vspace{-10pt}
\end{table}

\noindent{\textbf{Model Architecture.}} 
We study the effectiveness of the Mamba architecture, the image degradation prior (IDP) and the MoE Model. As shown in \cref{tab:ab1} , the best performance is achieved when all modules are used.

\begin{table}[ht]
\vspace{-2mm}
  \centering
  \caption{\small \it The effectiveness of our model components. 
  }
  \vspace{-10pt}
  \setlength{\tabcolsep}{4pt}
  \begin{tabular}{c|cccc|cc}
    \Xhline{4\arrayrulewidth} 
    Case & Baseline & Mamba & IDP & MoE & PSNR$\textcolor{black}{\uparrow}$ & SSIM$\textcolor{black}{\uparrow}$   \\
    \Xhline{4\arrayrulewidth}
    1 & \ding{51} & \ding{55} & \ding{55} & \ding{55} & 16.70 & 0.5779 \\
    2 & \ding{51} & \ding{51} & \ding{55} & \ding{55} & 17.24 & 0.6481 \\ 
    3 & \ding{51} & \ding{51} & \ding{51} & \ding{55} & 18.10 & 0.7539 \\      
    \textbf{Ours} & \ding{51} & \ding{51} & \ding{51} & \ding{51} & \textbf{18.49} & \textbf{0.7860} \\
    \Xhline{1.5pt} 
    \end{tabular}%
    \vspace{-10pt}
  \label{tab:ab1}%
\end{table}%

\noindent{\textbf{Downstream Validation.}} 
\begin{figure}[ht]
    \centering
    \begin{tikzpicture}
    \node[anchor=south west,inner sep=0] (image) at (0,0) {\includegraphics[width=\linewidth]{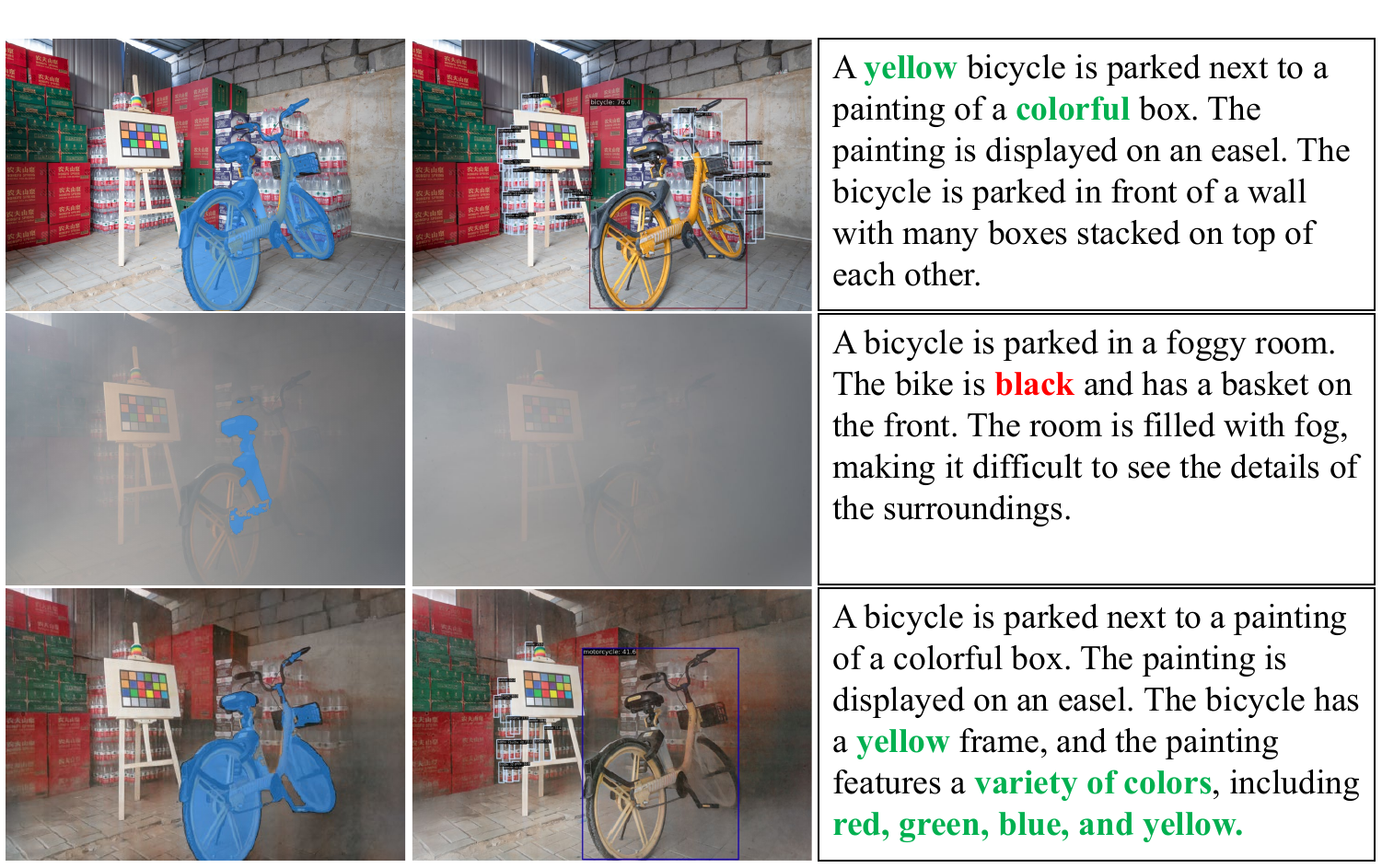}};
        \begin{scope}[x={(image.south east)},y={(image.north west)}]
            \draw (0.155,  0.99) node {\small Segmentation};
            \draw (0.435,  0.99) node {\small Detection};
            \draw (0.77,  0.99) node {\small Caption};
        \end{scope}
    \end{tikzpicture}
    \vspace{-18pt}
    \caption{\small \it Downstream task evaluation. We validate SAM \cite{SAM}, DEQDet \cite{DEQDet}, and LLaVA \cite{LLaVa} on three downstream tasks. From top to bottom, we show the results of GT, hazy images, and our dehazed images on three downstream tasks, respectively. \textcolor{myGreen}{Green} and \textcolor{red}{red} font colours indicate correct and incorrect descriptions. Best viewed in colour on the screen.}
    \Description{Downstream task evaluation.}
    \label{fig:downstream}
    \vspace{-5mm}
\end{figure}
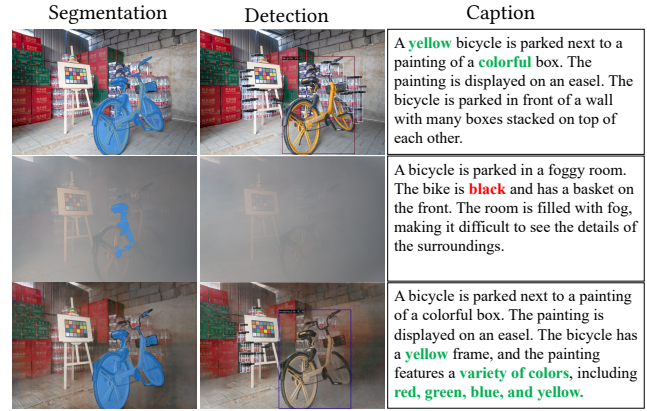
One goal of image dehazing is to serve downstream high-level vision tasks. Therefore, evaluating the performance of dehazing methods on these tasks becomes essential. Here, we select three typical tasks with associated SOTA methods as baselines: object detection with DEQDet \cite{DEQDet}, image segmentation with SAM \cite{SAM}, and image captioning with LLaVA \cite{LLaVa}. 

As shown in~\cref{fig:downstream}, current SOTA baselines still struggle under dense haze conditions, \eg, the undetected bicycle. When we input our dehazed image into the baselines for downstream tasks, we observe significant performance improvements. This finding underscores the essential role of image dehazing as a preprocessing step. For space reason, in the supplementary materials, we provide a quantitative comparison of various SOTA dehazing methods when used as pre-processing steps for downstream tasks. Notably, images dehazed using our MoE-Mamba framework exhibit improved performance across all three downstream tasks.

\section{Conclusion}
In this work, we propose a large-scale and real-world high quality dehazing dataset LMHaze. Our LMHaze provides high-resolution images with multiple haze intensities and rich semantic annotation. Moreover, we conduct an extensive benchmark using SOTA dehazing methods with an additional new metric based on a large multimodal model to simulate human perception for evaluating dehazed images. Meanwhile, to better handle image dehazing, we propose a new intensity-aware single-image dehazing method, MoE-Mamba. It dynamically adjusts the model parameters according to the input hazy image. Extensive experiments demonstrate the high quality of our LMHaze dataset, and our MoE-Mamba framework achieves competitive performance compared to SOTA methods.



\bibliographystyle{ACM-Reference-Format}
\bibliography{main}










\end{document}